\documentclass[10pt,twocolumn,letterpaper]{article}

\usepackage{cvpr}
\usepackage{times}
\usepackage{epsfig}
\usepackage{graphicx}
\usepackage{amsmath}
\usepackage{amssymb}
\usepackage{multirow}
\usepackage{stmaryrd,subfig,color}

\cvprfinalcopy 

\begin{document}

\title{Frequency-Tuned Universal Adversarial Attacks}

\author{Yingpeng Deng$^{1}$ and Lina J. Karam$^{1,2}$\\
$^{1}$Image, Video and Usability Lab, School of ECEE, Arizona State University, Tempe, AZ, USA\\
$^{2}$Dept. of ECE, School of Engineering, Lebanese American University, Lebanon\\
\tt\small{\{ypdeng,karam\}@asu.edu}}

\maketitle

\begin{abstract}

Researchers have shown that the predictions of a convolutional neural network (CNN) for an image set can be severely distorted by one single image-agnostic perturbation, or universal perturbation, usually with an empirically fixed threshold in the spatial domain to restrict its perceivability. However, by considering the human perception, we propose to adopt JND thresholds to guide the perceivability of universal adversarial perturbations. Based on this, we propose a frequency-tuned universal attack method to compute universal perturbations and show that our method can realize a good balance between perceivability and effectiveness in terms of fooling rate by adapting the perturbations to the local frequency content. Compared with existing universal adversarial attack techniques, our frequency-tuned attack method can achieve cutting-edge quantitative results. We demonstrate that our approach can significantly improve the performance of the baseline on both white-box and black-box attacks.

\end{abstract}

\section{Introduction}
\label{sec:intro}

\begin{figure*}[t]
    \centering
	\begin{minipage}{0.19\textwidth}
		\centering
		\normalsize{No attack}
	\end{minipage}
	\begin{minipage}{0.19\textwidth}
		\centering
		\normalsize{UAP~\cite{moosavi2017universal}}
	\end{minipage}
	\begin{minipage}{0.19\textwidth}
		\centering
		\normalsize{FTUAP-FF}
	\end{minipage}
	\begin{minipage}{0.19\textwidth}
		\centering
		\normalsize{FTUAP-MHF}
	\end{minipage}
	\begin{minipage}{0.19\textwidth}
		\centering
		\normalsize{FTUAP-MF}
	\end{minipage}
    \begin{minipage}{0.19\textwidth}
        \centering
        \includegraphics[width=0.9\textwidth]{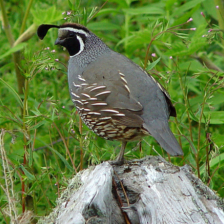}
	\end{minipage}
    \begin{minipage}{0.19\textwidth}
        \centering
        \includegraphics[width=0.9\textwidth]{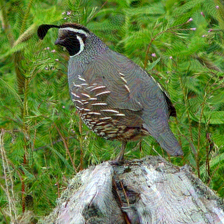}
    \end{minipage}
    \begin{minipage}{0.19\textwidth}
        \centering
        \includegraphics[width=0.9\textwidth]{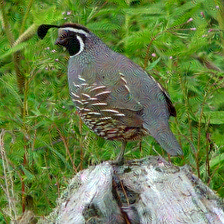}
    \end{minipage}
    \begin{minipage}{0.19\textwidth}
        \centering
        \includegraphics[width=0.9\textwidth]{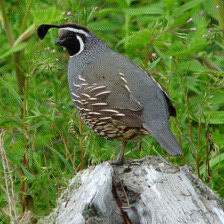}
    \end{minipage}
    \begin{minipage}{0.19\textwidth}
        \centering
        \includegraphics[width=0.9\textwidth]{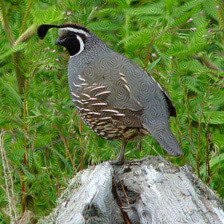}
    \end{minipage}
    \begin{minipage}{0.19\textwidth}
		\centering
		\textcolor[rgb]{0,0.5,0}{\normalsize{quail 100\%}}\\
	\end{minipage}
	\begin{minipage}{0.19\textwidth}
		\centering
		\textcolor[rgb]{0,0.5,0}{\normalsize{quail 97\%}}\\
	\end{minipage}
	\begin{minipage}{0.19\textwidth}
		\centering
		\textcolor[rgb]{0.7,0,0}{\normalsize{peacock 99\%}}\\
	\end{minipage}
	\begin{minipage}{0.19\textwidth}
		\centering
		\textcolor[rgb]{0.7,0,0}{\normalsize{peacock 89\%}}\\
	\end{minipage}
	\begin{minipage}{0.19\textwidth}
		\centering
		\textcolor[rgb]{0.7,0,0}{\normalsize{peacock 96\%}}\\
	\end{minipage}
    \begin{minipage}{0.19\textwidth}
        \centering
        \includegraphics[width=0.9\textwidth]{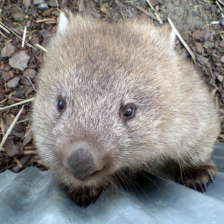}
	\end{minipage}
    \begin{minipage}{0.19\textwidth}
        \centering
        \includegraphics[width=0.9\textwidth]{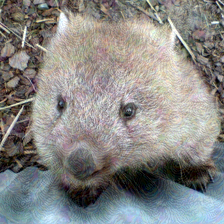}
    \end{minipage}
    \begin{minipage}{0.19\textwidth}
        \centering
        \includegraphics[width=0.9\textwidth]{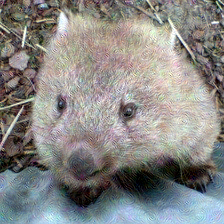}
    \end{minipage}
    \begin{minipage}{0.19\textwidth}
        \centering
        \includegraphics[width=0.9\textwidth]{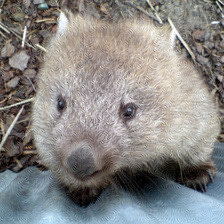}
    \end{minipage}
    \begin{minipage}{0.19\textwidth}
        \centering
        \includegraphics[width=0.9\textwidth]{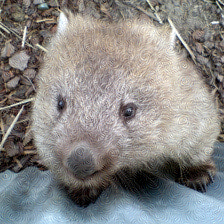}
    \end{minipage}
    \begin{minipage}{0.19\textwidth}
		\centering
		\textcolor[rgb]{0,0.5,0}{\scriptsize{wombat 99\%}}\\
	\end{minipage}
	\begin{minipage}{0.19\textwidth}
		\centering
		\textcolor[rgb]{0.7,0,0}{\normalsize{brain coral 47\%}}\\
	\end{minipage}
	\begin{minipage}{0.19\textwidth}
		\centering
		\textcolor[rgb]{0.7,0,0}{\normalsize{brain coral 52\%}}\\
	\end{minipage}
	\begin{minipage}{0.19\textwidth}
		\centering
		\textcolor[rgb]{0.7,0,0}{\normalsize{brain coral 84\%}}\\
	\end{minipage}
	\begin{minipage}{0.19\textwidth}
		\centering
		\textcolor[rgb]{0.7,0,0}{\normalsize{brain coral 87\%}}\\
	\end{minipage}
    \begin{minipage}{0.19\textwidth}
        \centering
        \includegraphics[width=0.9\textwidth]{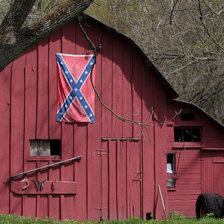}
	\end{minipage}
    \begin{minipage}{0.19\textwidth}
        \centering
        \includegraphics[width=0.9\textwidth]{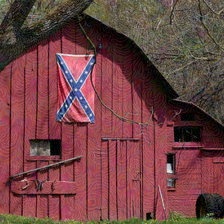}
    \end{minipage}
    \begin{minipage}{0.19\textwidth}
        \centering
        \includegraphics[width=0.9\textwidth]{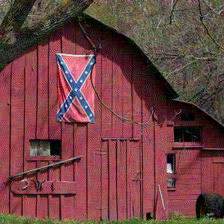}
    \end{minipage}
    \begin{minipage}{0.19\textwidth}
        \centering
        \includegraphics[width=0.9\textwidth]{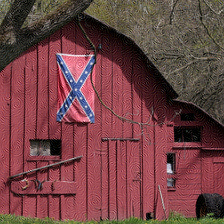}
    \end{minipage}
    \begin{minipage}{0.19\textwidth}
        \centering
        \includegraphics[width=0.9\textwidth]{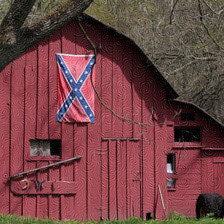}
    \end{minipage}
    \begin{minipage}{0.19\textwidth}
		\centering
		\textcolor[rgb]{0,0.5,0}{\normalsize{barn 100\%}}\\
	\end{minipage}
	\begin{minipage}{0.19\textwidth}
		\centering
		\textcolor[rgb]{0,0.5,0}{\normalsize{barn 73\%}}\\
	\end{minipage}
	\begin{minipage}{0.19\textwidth}
		\centering
		\textcolor[rgb]{0.7,0,0}{\normalsize{peacock 47\%}}\\
	\end{minipage}
	\begin{minipage}{0.19\textwidth}
		\centering
		\textcolor[rgb]{0.7,0,0}{\normalsize{backpack 12\%}}\\
	\end{minipage}
	\begin{minipage}{0.19\textwidth}
		\centering
		\textcolor[rgb]{0.7,0,0}{\normalsize{safety pin 40\%}}\\
	\end{minipage}
    \caption{Examples of perturbed images. From left to right: unperturbed image, perturbed images by UAP in the spatial domain~\cite{moosavi2017universal} and our frequency-tuned UAP (FTUAP) in all frequency bands (FF), middle and high frequency bands (MHF), and middle frequency bands (MF). The original (green) and perturbed predictions (red) with their corresponding confidence scores are listed under each image.}
    \label{fig:1}
\end{figure*}

Convolutional neural networks (CNNs) have shown great potential in various computer vision problems; however, these CNNs were shown to be vulnerable to perturbations in their input, even when such perturbations are small~\cite{szegedy2013intriguing}. By adding such a small perturbation to the input image, a CNN-based classifier can be easily fooled and can alter its predictions, while a human can still correctly classify the perturbed input image. Furthermore, researchers conducted various studies on the vulnerabilities of CNNs to adversarial attacks in the image understanding area, inlcuding image classification ~\cite{goodfellow2014explaining,moosavi2016deepfool,kurakin2016adversarial,moosavi2017universal} and semantic segmentation ~\cite{xie2017adversarial,xiao2018characterizing,poursaeed2018generative}. Generally, the proposed adversarial attack algorithms can be categorized as black-box attacks and white-box attacks according to the accessibility to the attacked model. Usually, a black-box attack have little or no access to the targeted model, with its predictions as the only possible feedback~\cite{bhambri2019survey}. Thus, one can perform such an attack by making use of gradient estimation~\cite{chen2017zoo,bhagoji2018practical,liu2019geometry}, transferability~\cite{papernot2017practical,moosavi2017universal}, local search~\cite{guo2019simple,li2019nattack} or combinatorics~\cite{moonICML19}.

On the contrary, for a white-box attack, we have full knowledge about the targeted model, such as its architecture, weight and gradient information, during the computation of the perturbation. In this case, the perturbation can be optimized effectively with the model information and an objective function by backpropagation. Additionally, there are thousands of publicly available CNN models on which we can carry out white-box adversarial attacks. Thus, to find a small but effective adversarial perturbation, we tend to utilize the white-box attack. One interesting direction of the white-box attack is the universal adversarial attack. In this latter case, a single perturbation, which is computed using a training set with a relatively small number of images, can be applied to input images to fool a targeted model~\cite{moosavi2017universal}. Given the good cross-model generalization of white-box attacks~\cite{moosavi2017universal,reddy2018nag}, we can also extend the universal perturbations that are pretrained on one or multiple popular CNN models to black-box attacks on other unseen models.

However, most of the universal adversarial attack methods attempt to decrease the perceptibility of the computed perturbation $\delta$ by limiting the $l_p$ norm, $p\in [1,\infty)$, to not exceed a fixed threshold (for instance, $\left\| \delta \right\|_\infty \leq 10$ for an 8-bit image). This type of thresholding method treats different image/texture regions identically and does not take human perception sensitivity into consideration. Aiming to solve this problem, we introduce the just-noticeable-difference (JND) into our universal attacks. JND is the minimum contrast for a human to perceive the change before and after adding such a small value. In fact, JND is not fixed but varies with the local image characteristics including but not limited to background intensity and frequency content. For example, human contrast sensitivity can be regarded as a function of frequency. Generally, a human is more sensitive to intensity changes that occur in the image regions that are dominated by low to middle frequency content, which means JNDs are expected to be lower in such regions~\cite{ahumada1992luminance,hontsch2002adaptive} than those with high-mid and high frequency content. Consequently, it is more reasonable to adaptively adjust the thresholds for different regions based on frequency content with the guidance of JND.

Therefore, instead of directly computing the perturbations directly in the spatial image, which has been widely used by many universal attack algorithms, we propose to conduct our attacks in the frequency domain by adopting the perception-based JND threshold and tuning the computed perturbation based on frequency content. Moreover, we can also examine the effects of different frequency bands on training perturbations as well as find a balance between the good effectiveness of universal attacks and the invisibility to human eyes. Some adversarial examples that are produced with our proposed frequency-tuned universal attack (FTUAP) method are shown in Figure~\ref{fig:1}.

Our main contributions are summarized as follows:
\begin{itemize}
  \item [1)] 
  To the best of our knowledge, we are the first ones to take human perception sensitivity into account to compute imperceptible or quasi-imperceptible universal perturbations. We propose to adopt a JND threshold to mimic the perception sensitivity of human vision and guide the computation of universal perturbations. The JND threshold for each frequency band is calculated specifically based on the parametric model which approximates luminance-based contrast sensitivity~\cite{ahumada1992luminance,hontsch2002adaptive}.
  \item [2)]
  We conduct a series of ablation experiments, on the universal perturbations trained in different frequency bands. We test and compare the results with respect to attack performance and visibility extensively and suggest that the universal attacks on middle and high frequency bands can be both effective and nearly imperceptible. 
  \item [3)]
  We show that the universal perturbation by our method can significantly improve the performance in terms of fooling rate/top-1 accuracy as compared to the normal universal attack method, achieve a similar or higher attacking result with much less perceptibility, and also present stronger ability even when attacking the defended models.
  \item[4)]
  By collecting histograms of perturbation values in both spatial and frequency domains, we analyze the perturbation distribution properties of our algorithm with repect to the difference from UAP~\cite{moosavi2017universal} and JND thresholds.
\end{itemize}

\section{Related Works}
\label{sec:works}

In this paper, we mainly focus on the white-box universal adversarial attacks given its light size (the same as one single image), considerable effectiveness (can cause severe malfunction of the targeted model on the image set) and promising transferability (good cross-model generalization). First, we will introduce some image-dependent attack algorithms because they can be adopted into the pipeline of some universal attack methods, followed by related work on universal attacks and frequency-based attacks.

\subsection{Image-dependent Attacks}
\label{ssec:imagedep}

The attack power can be maximized by computing an image-dependent perturbation for each input image separately. Goodfellow \textit{et al}~\cite{goodfellow2014explaining} found that only a single step of gradient ascent, referred to as fast gradient sign method (FGSM), based on the original input image and loss function can generate an adversarial example with a large fooling probability, within an almost imperceptible scale compared to the image scale. Moosavi-Dezfooli \textit{et al}~\cite{moosavi2016deepfool} proposed DeepFool to compute the minimum perturbation that reaches the decision boundary of the classifier. Later, Kurakin \textit{et al}~\cite{kurakin2016adversarial} introduced an iterative gradient sign method (IGSM) algorithm, which consists of a multistep FGSM. An alternative for gradient calculation is the least likely class method~\cite{kurakin2016adversarial}. Instead of maximizing the loss on the true label, it minimizes the loss on the label with the lowest prediction probability for the clean image. Modas \textit{et al}~\cite{modas2019sparsefool} showed that strong quasi-imperceptible attacks can be obtained by perturbing only few pixels without limiting the perturbation's norm.

\subsection{Universal Attacks}
\label{ssec:universal}

While the aforementioned methods deal with image-dependent adversarial perturbations, the generation of a universal, image-agnostic perturbation is a more challenging topic because it aims to find one universal perturbation that can drastically reduce the prediction accuracy when applied to any input image. Based on the DeepFool algorithm~\cite{moosavi2016deepfool}, Moosavi-Dezfooli \textit{et al}~\cite{moosavi2017universal} generated universal adversarial perturbations (UAP) by accumulating the updates iteratively for all the training images. Fast Feature Fool~\cite{mopuri-bmvc-2017} and GD-UAP~\cite{mopuri2018generalizable} did not make use of training data but rather aimed to produce a perturbation by inputing it into the targeted model and maximizing the mean activations of different hidden layers, while the latter also took advantage of image statistical information. The data-independent methods are unsupervised and not as strong as the aforementioned supervised ones. Thus, in this paper, we will mainly focus on supervised universal perturbations. Recently, generative adversarial networks were adopted to generate the universal perturbation in~\cite{poursaeed2018generative,reddy2018nag}, where the adversarial networks were set as the targeted models. Poursaeed \textit{et al}~\cite{poursaeed2018generative} trained the generative network using a fixed random pattern as input to produce the generative adversarial perturbation (GAP), while Mopuri \textit{et al}~\cite{reddy2018nag} introduced both fooling and diversity objectives as the loss function of their network for adversary generation (NAG) to learn the perturbation distribution through random patterns sampled from the latent space. The authors of~\cite{mummadi2019defending} and~\cite{shafahi2018universal} (iPGD) use mini-batch based stochastic PGD during training by maximizing the average loss over each mini-batch.

\subsection{Frequency Attacks}
\label{ssec:freq}

Prior to our work, there have been several papers performing adversarial attacks in the frequency domain. Guo \textit{et al}~\cite{guo2018low} presented a black-box attack by randomly searching in low frequency bands in the DCT domain. Sharma \textit{et al}~\cite{sharma2019effectiveness} showed that both undefended and defended models are vulnerable to low frequency attacks. However, the authors of~\cite{guo2018low,sharma2019effectiveness} set the limit on the perturbation norm in the spatial domain instead of the frequency domain, and these existing methods generate significantly perceivable perturbations by focusing on low frequency components and suppressing the high frequency components. The resulting low-frequency perturbations can be easily perceived by humans even with a small perturbation norm. Wang \textit{et al}~\cite{wang2019high} improved the robustness of CNNs against adversarial attacks by preserving the low frequency components of input images. Their attacks consist of blurring the input images by removing high frequency components, which is 
equivalent to low pass filtering the input images with the degree of the "blur" attack controlled by how many high frequency components get removed or filtered out.

\section{Frequency-Tuned Attacks}
\subsection{Discrete Cosine Transform}
\label{ssec:dct}

\subsubsection{Formula Representation}
\label{ssssec:formula}

The Discrete Cosine Transform (DCT) is a fundamental tool that is used in signal, image and video processing, especially for data compression, given its properties of energy compaction in the frequency domain and real-number transform. In our paper, we adopt the orthogonal Type-II DCT transform, whose formula is identical to its inverse DCT transform. The DCT transform formula can be expressed as:

\begin{equation}
    X(k_1,k_2)=\sum_{n_1=0}^{N_1-1}\sum_{n_2=0}^{N_2-1}x(n_1,n_2)c_1(n_1,k_1)c_2(n_2,k_2)
\label{eq1}
\end{equation}
\begin{equation}
\begin{aligned}
    c_i(n_i,k_i)=\tilde{c_i}(k_i)\cos\left(\frac{\pi(2n_i+1)k_i}{2N_i}\right),\\ \ 0\leq n_i, k_i\leq N_i-1, \ i=1, 2
\label{eq2}
\end{aligned}
\end{equation}
\begin{equation}
    \tilde{c_i}(k_i)=\left\{
    \begin{array}{rcl}
    \sqrt{\frac{1}{N_i}}, & k_i = 0\\
    \sqrt{\frac{2}{N_i}}, & k_i \neq 0
    \end{array}
    \right., \ i=1, 2.
\label{eq3}
\end{equation}
In Equation~\ref{eq1}, $x(n_1,n_2)$ is the image pixel value at location $(n_1,n_2)$ and $X(k_1,k_2)$ is the DCT transform of the $N_1\times N_2$ image block. Usually, we set $N_1 = N_2 = 8$. Using the DCT transform, an $8\times 8$ spatial block $x(n_1,n_2)$  can be converted to an $8\times 8$ frequency response block $X(k_1,k_2)$, with a total of 64 frequency bands. Suppose we have a $224\times 224$ gray image $x$. Given that the DCT transform is block-wise with $N_1 = N_2 = 8$,  we can divide the image into $28\times 28 =784$ non-overlapping $8\times 8$ blocks. Then we can obtain its DCT transform $X(\hat{n}_1, \hat{n}_2, k_1, k_2)$, where $k_1, k_2$ denote 2-D frequency indices and the 2-D block indices $(\hat{n}_1, \hat{n}_2)$ satisfy $0\leq \hat{n}_1, \hat{n}_2 \leq 27$.

\subsubsection{Matrix Representation}
\label{sssec:matrix}

For better computation efficiency, we can represent the DCT transform as a matrix operation. Here, let $c(n_1,n_2,k_1,k_2) = c_1(n_1,k_1)c_2(n_2,k_2)$, and further perform a flattening operation by representing the spatial indices $n_1,n_2$ as $s$ and frequency indices $k_1,k_2$ as $b$, where $s=8n_1+n_2$ and $b=8k_1+k_2$, $\ 0\leq s,b \leq 63$. We define $C=\{c(s,b)\}$ as the $64 \times 64$ DCT coefficient matrix, where $s$ corresponds to the flattened spatial index of the block and $b$ indicates the flattened frequency index. If we consider an $8 \times 8$ image block $x$ and its vectorized version $x_v$, the DCT transform can be written as:

\begin{equation}
    X_v = C^Tx_v,
\label{eq4}
\end{equation}
which is a simple matrix dot product operation and $X_v$ is the vectorized frequency response in the DCT domain. To compute the DCT transform for a $224 \times 224$ image, one can construct $x_v$ as a stack of multiple vectorized image blocks with proper reshaping. In this case, $x_v$ becomes a $784 \times 64$ image stack with each row corresponding to the vectorized $8 \times 8$ image block, and 784 indicates the number of image blocks, and $X_v$ is also a $784 \times 64$ stack with vectorized DCT transform blocks in each row. This can be easily extended to color images or image batches by adding channels to the stack. With this representation, we can directly plug the DCT transform in the backpropagation process to efficiently compute the perturbation within the frequency domain. In fact, by only flattening the frequency indices as $b$, we can directly add a convolutional layer with a fixed DCT matrix $C=\{c(n_1,n_2,b)\}$ and a stride of 8 to implement the block-wise DCT transform followed by inverse DCT transform before the targeted CNN model.

\subsection{Perception-based JND Thresholds}
\label{ssec:jnd}

Inspired by human perception sensitivity, we compute the JND thresholds in the DCT domain by adopting the luminance-model-based JND thresholds~\cite{ahumada1992luminance,hontsch2002adaptive}. Overall, the JND thresholds for different frequency bands can be computed as

\begin{equation}
    t_{DCT}(k_1,k_2) = \frac{MT(k_1,k_2)}{2\tilde{c_1}(k_1)\tilde{c_2}(k_2)(L_{max}-L_{min})},
\label{eq5}
\end{equation}
where $L_{min}$ and $L_{max}$ are the minimum and maximum display luminance, and $M=255$ for 8-bit image. To compute the background luminance-adjusted contrast sensitivity $T(k_1,k_2)$, Ahumada \textit{et al} proposed an approximating parametric model\footnote{$T(k_1,k_2)$ can be computed for any $k_1$, $k_2$ which satisfy $k_1k_2\neq0$ by this model, while $T(0,0)$ needs to be estimated as min($T(0,1)$, $T(1,0)$).}~\cite{ahumada1992luminance}:

\begin{equation}
\begin{aligned}
    \log_{10}(T(k_1,k_2))=\log_{10}\frac{T_{min}}{r+(1-r)\cos^2\theta(k_1,k_2)}\\
    +K(\log_{10}f(k_1,k_2)-\log_{10}f_{min})^2.
\label{eq6}
\end{aligned}
\end{equation}
The frequency $f(k_1,k_2)$ and its orientation $\theta(k_1,k_2)$ are given as follows:

\begin{equation}
\begin{aligned}
    f(k_1,k_2) = \frac{1}{2N_{DCT}}\sqrt{\frac{k_1^2}{w_x^2}+\frac{k_2^2}{w_y^2}}, \\
    \theta(k_1,k_2)=\arcsin\frac{2f(k_1,0)f(0,k_2)}{f^2(k_1,k_2)},
\label{eq7}
\end{aligned}
\end{equation}
and the luminance-dependent parameters are generated by the following equations:

\begin{equation}
\begin{aligned}
    T_{min}=\left\{
    \begin{array}{cc}
        \left(\frac{L}{L_T}\right)^{\alpha_T}\frac{L_T}{S_0},&L\leq L_T  \\
        \frac{L}{S_0}, &L>L_T
    \end{array}
    \right., \\
    f_{min}=\left\{
    \begin{array}{cc}
        f_0\left(\frac{L}{L_f}\right)^{\alpha_f},&L\leq L_f  \\
        f_0, &L>L_f
    \end{array}
    \right.,\\
    K=\left\{
    \begin{array}{cc}
        K_0\left(\frac{L}{L_K}\right)^{\alpha_K},&L\leq L_K  \\
        K_0, &L>L_K
    \end{array}
    \right..
\label{eq8}
\end{aligned}
\end{equation}
The values of constants in equations~\ref{eq6}-\ref{eq8} are $r=0.7$, $N_{DCT}=8$, $L_T=13.45$ cd/m$^2$, $S_0=94.7$, $\alpha_T=0.649$, $f_0=6.78$ cycles/degree, $\alpha_f=0.182$, $L_f=300$ cd/m$^2$, $K_0=3.125$, $\alpha_K=0.0706$, and $L_K=300$ cd/m$^2$. Given a viewing distance of 60 cm and a 31.5 pixels-per-cm (80 pixels-per-inch), the horizontal width/vertical height of a pixel ($w_x/w_y$) is 0.0303 degree of visual angle~\cite{liu2006jpeg2000}. In practice, for a measured luminance of $L_{min}=0$ cd/m$^2$ and $L_{max}=175$ cd/m$^2$ ,we use the luminance $L$ corresponding to the median value of image during computation as following~\cite{karam2011efficient}:

\begin{equation}
    L=L_{min}+128\frac{L_{max}-L_{min}}{M}
\label{eq9}
\end{equation}

\begin{figure}[t]
\centering
\includegraphics[width=0.46\textwidth]{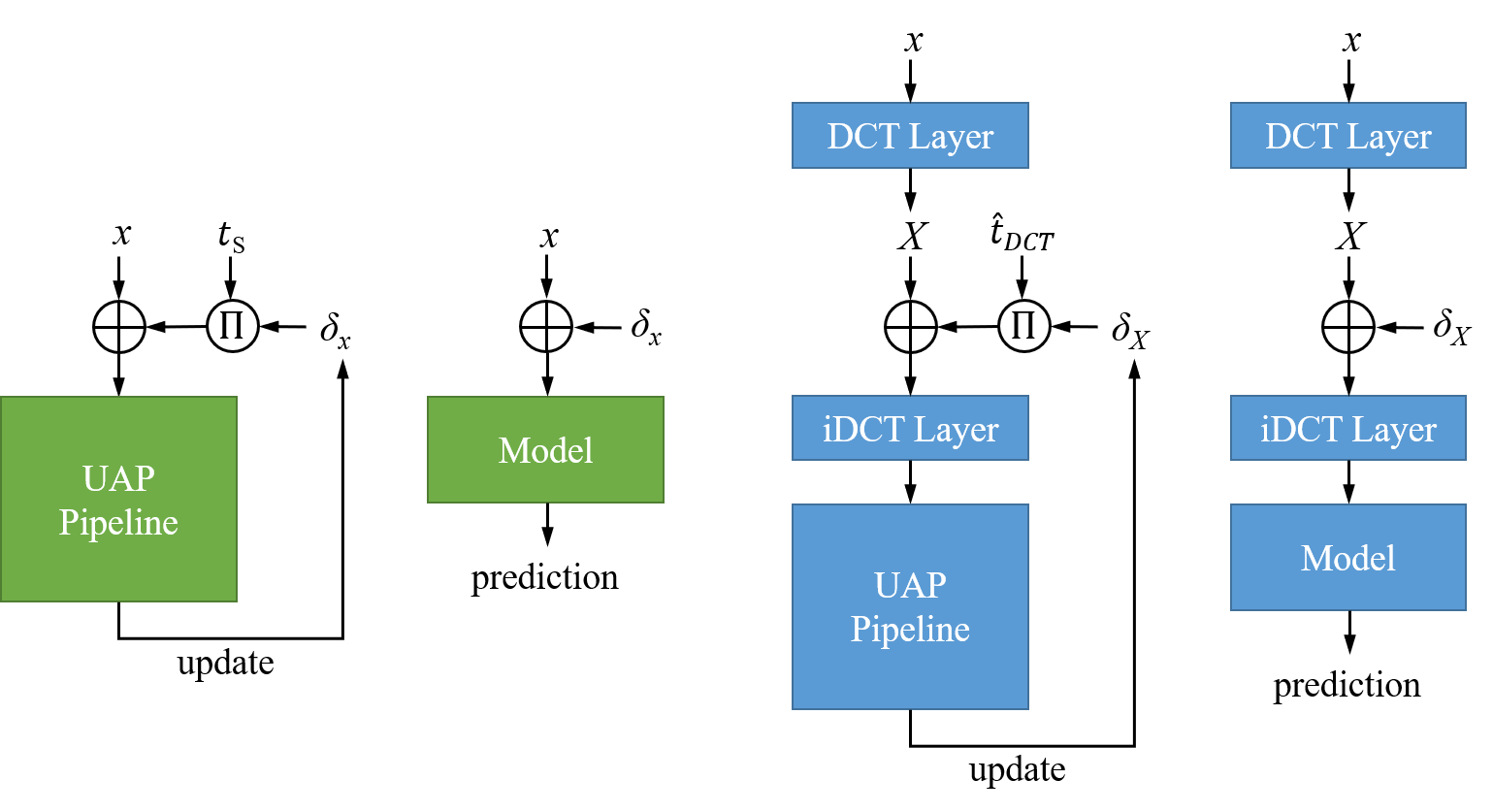}
	\begin{minipage}[t]{0.12\textwidth}
		\centering
		train
	\end{minipage}
    \begin{minipage}[t]{0.10\textwidth}
		\centering
		test
	\end{minipage}
	    \begin{minipage}[t]{0.14\textwidth}
		\centering
		train
	\end{minipage}
    \begin{minipage}[t]{0.10\textwidth}
		\centering
		test
	\end{minipage}
    \begin{minipage}{0.23\textwidth}
		\centering
		UAP
	\end{minipage}
	\begin{minipage}{0.23\textwidth}
		\centering
		FTUAP
	\end{minipage}
\caption{The flowchart examples for updating universal perturbations $\delta$ on the single image input by UAP and FTUAP. $\prod$ denotes the projection operation, i.e., thresholding in $l_\infty$ norm. $X$ indicates the DCT transform of the image $x$, while $\delta_{x}$ and $\delta_{X}$ are perturbations computed in the spatial domain and DCT domain, respectively.}
\label{fig:2}
\end{figure}

To be more flexible, we introduce a threshold coefficient $\lambda$. Thus, our final threshold $\hat{t}_{DCT} = \lambda t_{DCT}$. When $\lambda=2$, the JND threshold matrix $\hat{t}_{DCT}(k_1,k_2)$ for an 8-bit image in DCT domain is

\begin{equation}
\left[\tiny
\begin{array}{cccccccc}
    34.61&24.48&8.39&7.81&9.52&12.78&17.83&25.20\\
    24.48&12.47&6.91&6.32&7.45&9.76&13.39&18.71\\
    8.39&6.91&7.77&8.22&9.50&11.95&15.80&21.44\\
    7.81&6.32&8.22&10.27&12.49&15.53&19.93&26.20\\
    9.52&7.45&9.50&12.49&16.03&20.27&25.76&33.16\\
    12.78&9.76&11.95&15.53&20.27&26.09&33.28&42.44\\
    17.83&13.39&15.80&19.93&25.76&33.28&42.60&54.19\\
    25.20&18.71&21.44&26.20&33.16&42.44&54.19&68.78\\
\end{array}
\right].
\label{eq10}
\end{equation}
Also, we can train the perturbations on some specified frequency bands by partially setting $\hat{t}_{DCT} = 0$ on the irrelevant bands.

\subsection{Frequency-Tuned Universal Perturbations}
\label{ssec:ftattacks}

With the above matrix operation of DCT transform and JND threshold matrix, we can extend any existing gradient-based adversarial attack algorithm to a frequency-tuned attack. In our implementation, we adapt UAP~\cite{moosavi2017universal} to our frequency-tuned attacks. Figure~\ref{fig:2} illustrates the proposed frequency-tuned UAP (FTUAP) as compared to the spatial-domian UAP~\cite{moosavi2017universal}.

\begin{figure}[t]
\centering
\includegraphics[width=0.46\textwidth]{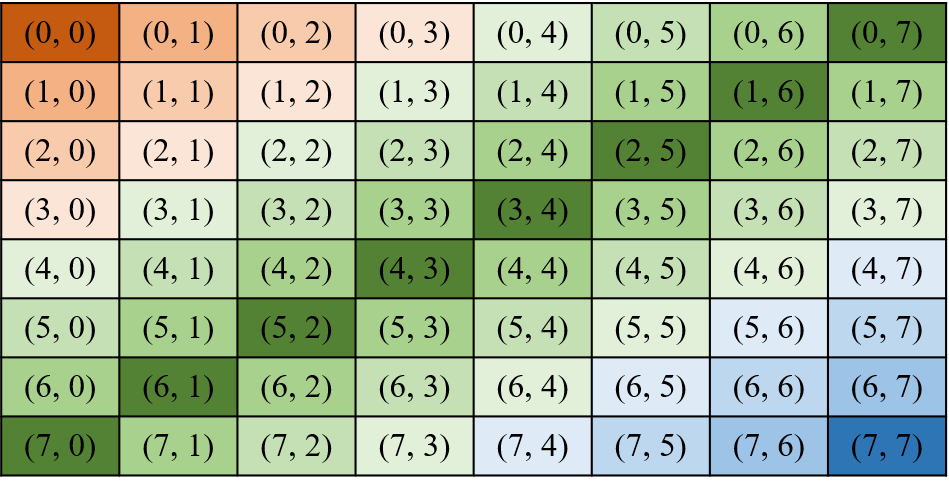}
\caption{The division of frequency. The location in each box indicates the 2-D frequency index $(k_1,k_2)$. The adjacent frequencies with the same color along diagonals are in the same slanting frequency bands. Low, middle and high frequency regions are presented in orange, green and blue (regardless of the color saturation).}
\label{fig:3}
\end{figure}

Given the $8 \times 8$ DCT frequency matrix, we recombine them as 15 nonoverlapping slanting frequency bands along the diagonal direction, i.e., the $i$-th slanting bands consists of all the previous frequency bands in $(k_1,k_2)$ such that $k_1+k_2=i, 0\leq i\leq 14$. Further, we divide the whole frequency bands as low ($0\leq i\leq 3$), middle ($4\leq i\leq 10$) and high ($11\leq i\leq 14$) frequency regions  without overlap, as illustrated in Figure~\ref{fig:3}.

\section{Experiments and Analyses}
\label{sec:experiments}

We compute the universal perturbations on 10000 randomly sampled images from the ILSVRC 2012 ImageNet~\cite{ILSVRC15} training set if not specified, and all of our implemented results (fooling rate/top-1 accuracy) are reported through evaluation on the full validation set with 50000 images of ImageNet. To restrict the perturbations during training, we set $\left\| \delta \right\|_\infty \leq 10$ for UAP and take (\ref{eq10}) as the DCT thresholds for our FTUAP. We stop the training process of the universal perturbations after 5 epochs on VGG and ResNet models, and after 10 epochs on Inception models. For quantative results, we adopt commonly used top-1 accuracy and fooling rate as the metrics. Given an image set $X$ and a universal perturbation $\delta$ in the spatial domain, the fooling rate can be expressed as $FR=\frac{1}{n}\sum_{i=1}^{n}\llbracket\hat{k}(x_i+\delta)\neq\hat{k}(x_i)\rrbracket, \quad x_i\in X$, where $\hat{k}(x_i)$ is the predicted label by the classifier.

\subsection{Perceivability vs. Effectiveness}
\label{ssec:perc}

\begin{figure*}[t]
    \centering
	\begin{minipage}{0.18\textwidth}
		\centering
        \includegraphics[width=1\textwidth]{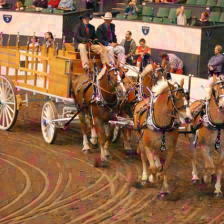}
	\end{minipage}
	\begin{minipage}[t]{0.1\textwidth}
		\centering
        \includegraphics[width=1\textwidth]{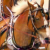}
	\end{minipage}
	\begin{minipage}{0.18\textwidth}
		\centering
        \includegraphics[width=1\textwidth]{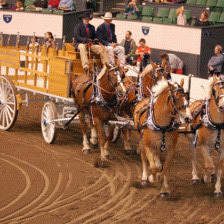}
	\end{minipage}
	\begin{minipage}[t]{0.1\textwidth}
		\centering
        \includegraphics[width=1\textwidth]{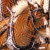}
	\end{minipage}
	\begin{minipage}{0.18\textwidth}
		\centering
        \includegraphics[width=1\textwidth]{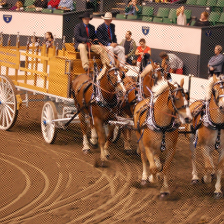}
	\end{minipage}
	\begin{minipage}[t]{0.1\textwidth}
		\centering
        \includegraphics[width=1\textwidth]{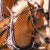}
	\end{minipage}
	\begin{minipage}{0.3\textwidth}
		\centering
        low freq.: 30.9\%
	\end{minipage}
	\begin{minipage}[t]{0.3\textwidth}
		\centering
        mid freq.: 84.9\%
	\end{minipage}
	\begin{minipage}{0.3\textwidth}
		\centering
        high freq.: 71.6\%
	\end{minipage}
	\begin{minipage}{0.18\textwidth}
		\centering
        \includegraphics[width=1\textwidth]{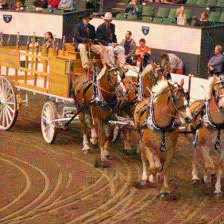}
	\end{minipage}
	\begin{minipage}[t]{0.1\textwidth}
		\centering
        \includegraphics[width=1\textwidth]{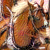}
	\end{minipage}
	\begin{minipage}{0.18\textwidth}
		\centering
        \includegraphics[width=1\textwidth]{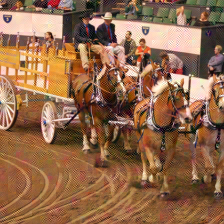}
	\end{minipage}
	\begin{minipage}[t]{0.1\textwidth}
		\centering
        \includegraphics[width=1\textwidth]{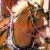}
	\end{minipage}
	\begin{minipage}{0.18\textwidth}
		\centering
        \includegraphics[width=1\textwidth]{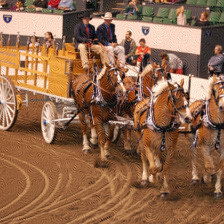}
	\end{minipage}
	\begin{minipage}[t]{0.1\textwidth}
		\centering
        \includegraphics[width=1\textwidth]{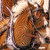}
	\end{minipage}
	\begin{minipage}{0.3\textwidth}
		\centering
        low\&mid freq.: 91.6\%
	\end{minipage}
	\begin{minipage}{0.3\textwidth}
		\centering
        low\&high freq.: 80.1\%
	\end{minipage}
	\begin{minipage}{0.3\textwidth}
		\centering
        mid\&high freq.: 91.7\%
	\end{minipage}
	\begin{minipage}{0.18\textwidth}
		\centering
        \includegraphics[width=1\textwidth]{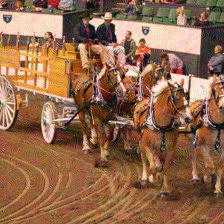}
	\end{minipage}
	\begin{minipage}[t]{0.1\textwidth}
		\centering
        \includegraphics[width=1\textwidth]{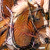}
	\end{minipage}
	\begin{minipage}{0.18\textwidth}
		\centering
        \includegraphics[width=1\textwidth]{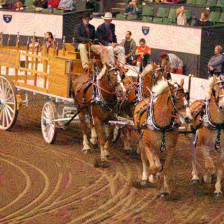}
	\end{minipage}
	\begin{minipage}[t]{0.1\textwidth}
		\centering
        \includegraphics[width=1\textwidth]{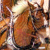}
	\end{minipage}
	\begin{minipage}{0.18\textwidth}
		\centering
        \includegraphics[width=1\textwidth]{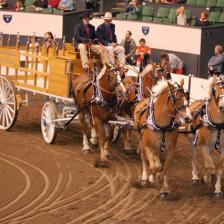}
	\end{minipage}
	\begin{minipage}[t]{0.1\textwidth}
		\centering
        \includegraphics[width=1\textwidth]{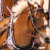}
	\end{minipage}
	\begin{minipage}{0.3\textwidth}
		\centering
        full freq.: 93.6\%
	\end{minipage}
	\begin{minipage}{0.3\textwidth}
		\centering
        spatial: 84.2\%
	\end{minipage}
	\begin{minipage}{0.3\textwidth}
		\centering
        no attack: 0\%
	\end{minipage}
    \caption{The visual examples of UAP in the spatial domain and our FTUAP in various combinations of frequency bands. For each example, we show the perturbed image (left) and its close-up (right), with corresponding attacking domain and fooling rate on validation set below the images.}
    \label{fig:4}
\end{figure*}

To show perceivability and effectiveness on different frequency bands, we use our FTUAP algorithm to train the perturbations on low, middle, high frequency bands and all the combined frequency bands (full frequency), separately, then we compare the visual examples and give the fooling rates evaluated on the whole validation set with those by UAP and unperturbed ones, as shown in Figure~\ref{fig:4}. For the FTUAP method, all the visual results are perturbed in the frequency domain and then inversely transformed to the spatial domain.

According to the first row in Figure~\ref{fig:4}, the low frequency perturbation generally causes only color artifacts (on the ground region in the image), and middle and high frequency perturbations produce more imperceptible texture patterns, for which we can observe the details in the close-ups. The perturbation in middle frequency bands contains more spiral patterns while high frequency patterns are visually more sparse yet locally intense. In the examples in the combinations of frequency bands, it seems that the patterns in different frequency bands are additive and can coexist. As compared to UAP, our FTUAP on full frequency bands shows similar perceivability on the whole perturbed image and we can only recognize stronger perturbation texture patterns in the close-ups, which means humans are more sensitive to the difference with low frequency components under the restriction of JND thresholds.

In terms of fooling rates, FTUAP on middle frequency bands can perform better than that on low or high frequency bands with a large gap, even exceeding UAP by 0.7\%. Further, FTUAP can achieve very high fooling rate results when combining middle frequency bands with others, especially 91.7\% of fooling rate on middle and high frequencies, which is 7.5\% higher but with a nearly imperceptible perturbation when compared to the perceivable UAP (spatial). With a similar perceivability, FTUAP on full frequency increases the fooling rate by about 10\% as compared to UAP. Generally, FTUAP gives higher fooling rates if trained on more frequency bands and produces the highest fooling rate after being trained on the full frequency bands.

\begin{table*}[t]
\centering
\caption{Comparison of fooling rates for different universal perturbations on pretrained CNN classifiers. Green indicates the best performance and blue indicates the second best performance.}
\label{comp1}
\begin{tabular}{|c|c|c|c|c|c|c|}
\hline
&VGG16&VGG19&ResNet50&ResNet152&GoogLeNet&Inception3\\
\hline
UAP~\cite{moosavi2017universal}&84.8&86.6&84.2&77.0&69.4&55.1\\
\hline
GAP~\cite{poursaeed2018generative}&83.7&80.1&-&-&-&\textcolor[rgb]{0,0.9,0}{82.7}\\
\hline
NAG~\cite{reddy2018nag}&77.6&83.8&86.6&87.2&\textcolor[rgb]{0,0.9,0}{90.4}&-\\
\hline
FTUAP-MF&86.4&86.0&84.9&83.1&72.8&60.9\\
FTUAP-MHF&\textcolor[rgb]{0,0,1}{90.1}&\textcolor[rgb]{0,0,1}{90.3}&\textcolor[rgb]{0,0,1}{91.7}&\textcolor[rgb]{0,0,1}{90.2}&82.3&\textcolor[rgb]{0,0,1}{71.3}\\
FTUAP-FF&\textcolor[rgb]{0,0.9,0}{93.5}&\textcolor[rgb]{0,0.9,0}{94.5}&\textcolor[rgb]{0,0.9,0}{93.6}&\textcolor[rgb]{0,0.9,0}{92.7}&\textcolor[rgb]{0,0,1}{85.8}&\textcolor[rgb]{0,0.9,0}{82.7}\\
\hline
\end{tabular}
\end{table*}

\begin{table*}[t]
\centering
\caption{Comparison of top-1 accuracy between iPGD attack method~\cite{shafahi2018universal} and our FTUAP-FF on 5000 training samples. Bold values correspond to best performance.}
\label{comp2}
\begin{tabular}{|c|c|c|c|c|c|}
\hline
&VGG16&ResNet152&GoogLeNet&Inceptionv3&mean\\
\hline
iPGD~\cite{shafahi2018universal}&22.5&16.4&\textbf{19.8}&\textbf{20.1}&19.7\\
\hline
FTUAP-FF&\textbf{9.3}&\textbf{12.0}&20.9&23.8&\textbf{16.5}\\
\hline
\end{tabular}
\end{table*}

\subsection{Fooling Capability}
\label{ssec:fool}

\begin{table*}[t]
\centering
\caption{Cross-model fooling rates. The first row displays the attacked models, and the first column indicates the targeted models for which adversarial perturbations were computed.}
\label{cross}
\begin{tabular}{|c|c|c|c|c|c|c|c|}
\hline
&&VGG16&VGG19&ResNet50&ResNet152&GoogLeNet&mean\\
\hline
\multirow{4}*{VGG16}&UAP&84.8&72.4&43.1&33.5&40.2&54.8\\
&FTUAP-MF&86.4&76.4&50.4&39.5&47.4&60.0\\
&FTUAP-MHF&90.1&79.9&54.0&40.5&46.3&62.2\\
&FTUAP-FF&93.5&84.7&50.3&38.3&42.4&61.8\\
\hline
\multirow{4}*{VGG19}&UAP&76.8&86.6&41.7&32.9&39.7&55.5\\
&FTUAP-MF&79.6&86.0&47.1&37.4&48.4&59.7\\
&FTUAP-MHF&82.1&90.3&50.8&39.6&49.5&62.5\\
&FTUAP-FF&87.0&94.5&47.7&36.8&42.3&61.7\\
\hline
\multirow{4}*{ResNet50}&UAP&64.9&61.0&84.2&45.3&45.0&60.1\\
&FTUAP-MF&73.2&64.6&84.9&50.3&47.8&64.2\\
&FTUAP-MHF&75.6&68.0&91.7&53.8&52.1&68.2\\
&FTUAP-FF&74.9&70.0&93.6&57.8&52.0&69.7\\
\hline
\multirow{4}*{ResNet152}&UAP&58.9&55.8&55.1&77.0&39.4&57.2\\
&FTUAP-MF&66.1&58.8&60.8&83.1&46.2&63.0\\
&FTUAP-MHF&79.3&71.4&75.3&90.2&59.0&75.0\\
&FTUAP-FF&74.5&69.7&77.1&92.7&56.0&74.0\\
\hline
\multirow{4}*{GoogLeNet}&UAP&57.9&56.8&43.7&35.0&69.4&52.6\\
&FTUAP-MF&66.2&64.3&49.1&42.0&72.8&58.9\\
&FTUAP-MHF&68.6&67.3&54.4&42.6&82.3&63.0\\
&FTUAP-FF&64.0&63.7&52.9&42.7&85.8&61.8\\
\hline
\end{tabular}
\end{table*}

To show the attacking ability of our FTUAP, we implement UAP and our FTUAP in middle frequency (FTUAP-MF), middle\&high frequency (FTUAP-MHF) and full frequency (FTUAP-FF) separately, on several modern CNN classifiers, including VGG~\cite{Simonyan15}, Inception\footnote{To implement block-wise DCT transform, the image size must be an integral multiple of 8, thus for Inception3 model, the $299 \times 299$ input image is resized to $296 \times 296$ for FTUAP.}~\cite{szegedy2015going,szegedy2016rethinking} and ResNet~\cite{he2016deep} classifiers pretrained on the ImageNet in the Pytorch library. Then we also compare our results with the published results of some state-of-the-art universal adversarial attack algorithms - GAP~\cite{poursaeed2018generative}, NAG~\cite{reddy2018nag} and iPGD~\cite{shafahi2018universal}, as listed in Tables~\ref{comp1} and~\ref{comp2}.

In Table~\ref{comp1}, our FTUAP-FF achieves the best performance in terms of fooling rate on most targeted models, and our FTUAP-MHF provides significantly less-perceivable perturbations while also performing well, especially on VGG and ResNet models. It is worth noting that by only attacking the middle frequency bands, FFTUAP-MF is able to result in similar or higher fooling rates on all the listed models as compared to UAP. FFTUAP-FF, which produces the best quantitative results among all the FFTUAP variants, significantly upgrades the attack performance of UAP by about 8\% on VGG16, VGG19 and ResNet50 models, more than 15\% on ResNet152 and GoogLeNet models, and over 20\% on the Inception3 model. From Table~\ref{comp2}, it can be seen that our FTUAP-FF exhibits a better overall attack power on the four models, specifically with lower top-1 accuracies on the VGG16 and ResNet152 networks and marginally higher ones on the GoogLeNet and Inception3 architectures. One can show that our frequency-tuned method would also enhance the iPGD algorithm by further reducing the top-1 accuracies, given the promising improvement from UAP to FTUAP.

On the other hand, we use perturbations which are computed for the specific targeted model to attack other untargeted models, to test if transferability can also be improved by our frequency-tuned algorithm, which serves as a very useful property for black-box attacks. Table~\ref{cross} shows that our method can not only equip UAP with stronger white-box attack ability, but also promote the cross-model generalization of the perturbation, i.e., stronger black-box attacks. For the generalization for each perturbation, UAP hardly produces a mean fooling rate of over 60\%, while all the perturbations by our FTUAP-MHF and FTUAP-FF result approximately in an overall increase by 7\%. In particular, FTUAP-MHF and FTUAP-FF reach 75\% and 74\%, respectively, in terms of mean fooling rate on ResNet152, whereas the highest one for UAP is only 60.1\% on ResNet50. Interestingly, according to the mean fooling rates, FTUAP-MHF generally shows better generalization on various models than FTUAP-FF.

\subsection{Attacks on Defended Models}
\label{defend}

\begin{table}[t]
\centering
\caption{Comparison of top-1 accuracy between UAP and FTUAP-FF by attacking the defended models in~\cite{borkar2019defending}.}
\label{robust}
\begin{tabular}{|c|c|c|c|}
\hline
&VGG16&ResNet152&GoogLeNet\\
\hline
No attack&66.3&79.0&67.8\\
\hline
UAP&62.5&76.7&65.5\\
\hline
FTUAP-FF&51.1&74.2&58.4\\
\hline
\end{tabular}
\end{table}

\begin{figure*}[t]
    \centering
	\begin{minipage}{0.24\textwidth}
		\centering
        \includegraphics[width=1\textwidth]{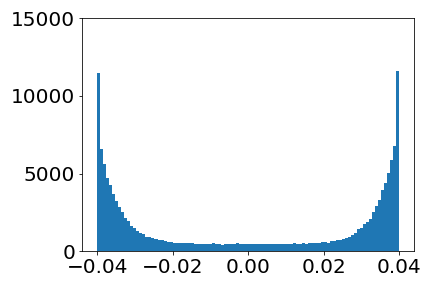}
	\end{minipage}
	\begin{minipage}{0.24\textwidth}
		\centering
        \includegraphics[width=1\textwidth]{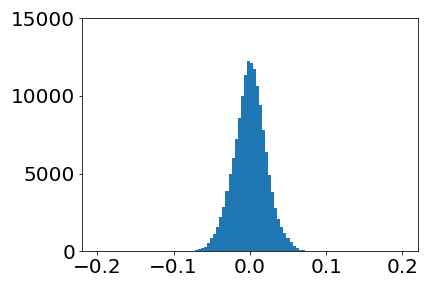}
	\end{minipage}
	\begin{minipage}{0.24\textwidth}
		\centering
        \includegraphics[width=1\textwidth]{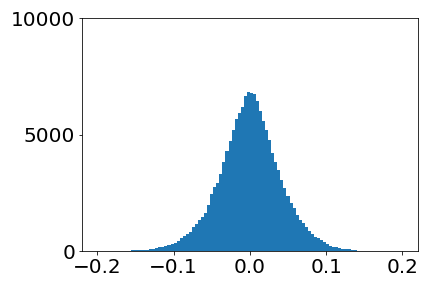}
	\end{minipage}
	\begin{minipage}{0.24\textwidth}
		\centering
        \includegraphics[width=1\textwidth]{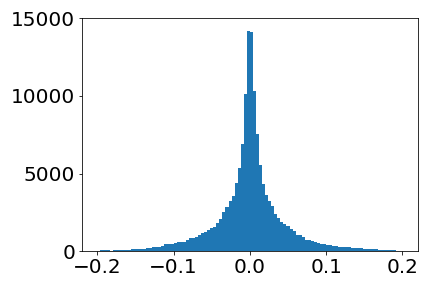}
	\end{minipage}
	\begin{minipage}{0.24\textwidth}
		\centering
        \normalsize{UAP: 0.0325}
	\end{minipage}
	\begin{minipage}{0.24\textwidth}
		\centering
        \normalsize{FTUAP-LF: 0.0210}
	\end{minipage}
	\begin{minipage}{0.24\textwidth}
		\centering
        \normalsize{FTUAP-MF: 0.0410}
	\end{minipage}
	\begin{minipage}{0.24\textwidth}
		\centering
        \normalsize{FTUAP-HF: 0.0463}
	\end{minipage}
	\begin{minipage}{0.24\textwidth}
		\centering
        \includegraphics[width=1\textwidth]{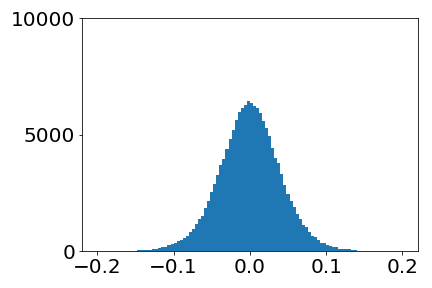}
	\end{minipage}
	\begin{minipage}{0.24\textwidth}
		\centering
        \includegraphics[width=1\textwidth]{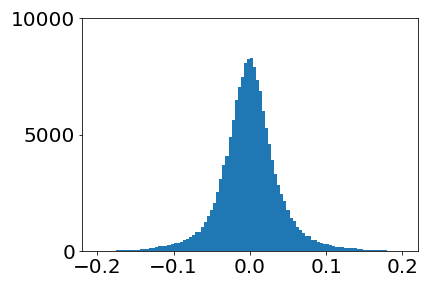}
	\end{minipage}
	\begin{minipage}{0.24\textwidth}
		\centering
        \includegraphics[width=1\textwidth]{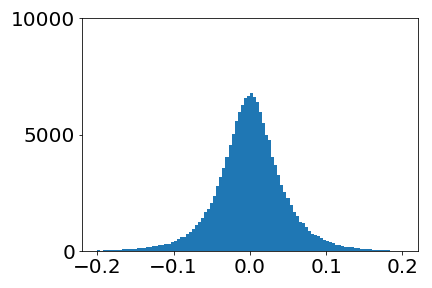}
	\end{minipage}
	\begin{minipage}{0.24\textwidth}
		\centering
        \includegraphics[width=1\textwidth]{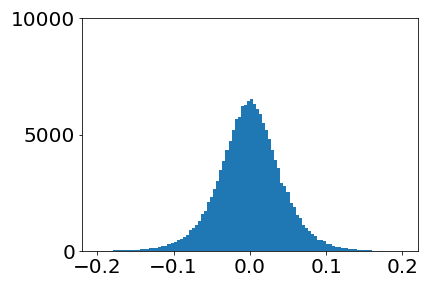}
	\end{minipage}
	\begin{minipage}{0.24\textwidth}
		\centering
        \normalsize{FTUAP-LMF: 0.0406}
	\end{minipage}
	\begin{minipage}{0.24\textwidth}
		\centering
        \normalsize{FTUAP-LHF: 0.0404}
	\end{minipage}
	\begin{minipage}{0.24\textwidth}
		\centering
        \normalsize{FTUAP-MHF: 0.0472}
	\end{minipage}
	\begin{minipage}{0.24\textwidth}
		\centering
        \normalsize{FTUAP-FF: 0.0437}
	\end{minipage}
    \caption{Histograms of different perturbations in the spatial domain. From left to right: (top) results by UAP, FTUAP in low, middle and high frequency (LF, MF, HF); (bottom) results by FTUAP in low\&middle, low\&high, mid\&high and full frequency (LMF, LHF, MHF, FF). Each histogram only describes one specific perturbation, under which the type of attack and the standard deviation of perturbation values are given. The image scale is normalized and all the perturbations by FTUAP have been transformed back to the spatial domain before visualization.}
    \label{fig:5}
\end{figure*}

Given that a number of advanced defense algorithms have been published against adversarial attacks, we also consider examining the potential of our FTUAP algorithm against the defended models. We adopt the currently strongest defense algorithm~\cite{borkar2019defending}. According to~\cite{borkar2019defending}, the defended model can outperform existing state-of-the art defense strategies and even effectively withstand unseen attacks via resilient feature regeneration. According to Table~\ref{robust}, our FTUAP-FF outperforms UAP on all three tested models. Altough UAP can only attack the defended models marginally with the top-1 accuracy decrease of less then 4\%, FTUAP-FF undermines the defense by reducing the accuracy with approximately 15\% and 9\% on VGG16 and GoogLeNet, respectively, which indicates that our method can also reinforce the attacking ability of UAP towards the defended models.

\subsection{Distributions of Perturbation Values}
\label{distribution}

We visualize the distributions of different FTUAP perturbations in the spatial domain together with a UAP perturbation in Figure~\ref{fig:5}. For the histogram of UAP, we find that the perturbation values are inclined to concentrate near the perturbation boundary ($\left\| \delta \right\|_\infty = 10$, i.e., $\left\| \delta \right\|_\infty \approx 0.04$ in the normalized image scale [0, 1].) Since FTUAP is not restricted directly in the spatial domain but rather in the frequency domain, the shown distributions are not obviously bounded. For the low frequency attack, its perturbation values cluster near zero with a narrow range, while the perturbation histogram of the middle frequency attack has a wider shape, meaning that higher frequency components lead to a wider dynamic range of perturbation values. When the frequency goes high, the histogram shows a Laplacian-like distribution with a sharp spike as well as longer tails. When more frequency bands are included in generating a FTUAP attack (bottom row of Figure~\ref{fig:5}), the distribution turns out to be Gaussian-like, which is similar for the attack which includes components in the middle frequency.

\begin{figure}[t]
    \centering
	\begin{minipage}{0.15\textwidth}
		\centering
        \includegraphics[width=1\textwidth]{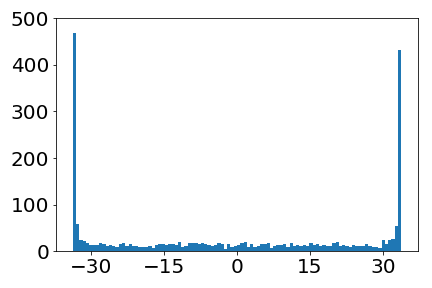}
	\end{minipage}
	\begin{minipage}{0.15\textwidth}
		\centering
        \includegraphics[width=1\textwidth]{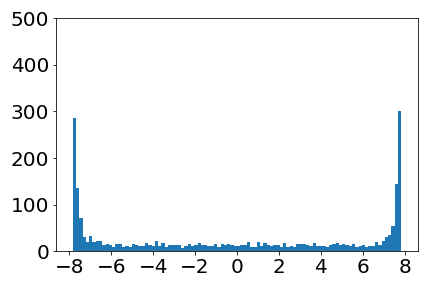}
	\end{minipage}
	\begin{minipage}{0.15\textwidth}
		\centering
        \includegraphics[width=1\textwidth]{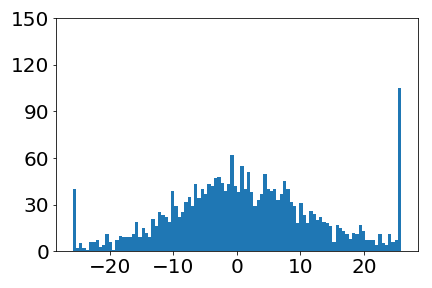}
	\end{minipage}
	\begin{minipage}{0.15\textwidth}
		\centering
        freq. (0, 0)
	\end{minipage}
	\begin{minipage}{0.15\textwidth}
		\centering
        freq. (0, 3)
	\end{minipage}
	\begin{minipage}{0.15\textwidth}
		\centering
        freq. (0, 7)
	\end{minipage}
	\begin{minipage}{0.15\textwidth}
		\centering
        \includegraphics[width=1\textwidth]{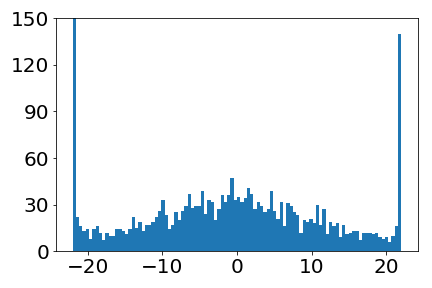}
	\end{minipage}
	\begin{minipage}{0.15\textwidth}
		\centering
        \includegraphics[width=1\textwidth]{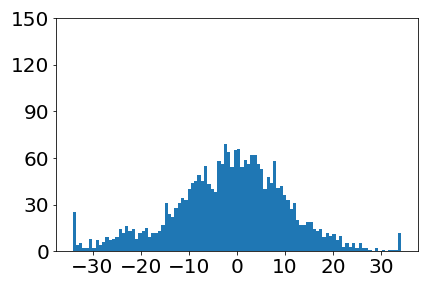}
	\end{minipage}
	\begin{minipage}{0.15\textwidth}
		\centering
        \includegraphics[width=1\textwidth]{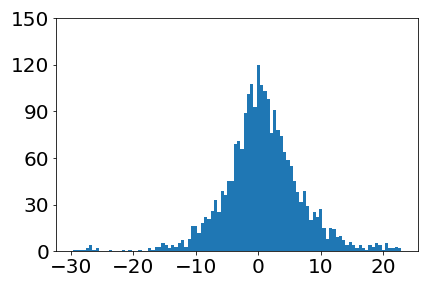}
	\end{minipage}
	\begin{minipage}{0.15\textwidth}
		\centering
        freq. (2, 7)
	\end{minipage}
	\begin{minipage}{0.15\textwidth}
		\centering
        freq. (4, 7)
	\end{minipage}
	\begin{minipage}{0.15\textwidth}
		\centering
        freq. (7, 7)
	\end{minipage}
    \caption{Histogram examples of perturbation distribution in one DCT frequency band. Below each histogram is the corresponding frequency band index as shown in Figure~\ref{fig:3}.}
    \label{fig:6}
\end{figure}

As our FTUAP computes the perturbations in the DCT domain, we show some histograms by extracting single frequency bands of perturbations trained by FTUAP-FF. According to Figure~\ref{fig:6}, many values are on the JND threshold boundary (i.e., abrupt high values at the two ends of the histograms for freq. (0, 0) and (0, 3)), which are very likely to be clipped by thresholds during training. When the frequency becomes higher, the values on the boundary starts to decrease and shift to zero. Given this, we infer that with the constraints of JND threshold, our FTUAP-FF can find the local optimal values more likely in middle and high frequency bands, while some low frequency components have their optimal solutions beyond the JND thresholds and thus suffer from truncation.

\section{Conclusion}
\label{conclusion}

Motivated by the fact that the human contrast sensitivity varies in functions of frequency, in our proposed frequency-tuned attack method, we integrate contrast sensitivity JND thresholds for generating frequency-domain perturbations that are tuned to the two-dimensional DCT frequency bands. In comparison with the baseline UAP, our FTUAP is able to achieve much higher fooling rates in a similar or even more imperceptible manner. Additionally, according to the conducted experiments, the proposed FTUAP significantly improves the universal attack performance as compared to existing universal attacks on various fronts including an increased fooling rates of  white-box attacks towards the targeted models, cross-model transferability for black-box attacks and attack power against defended models. The perturbation by our FTUAP method is adaptively bounded in the frequency domain as compared to existing method, and is more likely to reach the local optimum for middle and high frequency components.

\textbf{Acknowledgement.} We would like to thank Tejas Borkar for evaluating the perturbations on his defended models and providing the results in Table~\ref{robust}.

{\small
\bibliographystyle{ieee_fullname}
\bibliography{egbib}
}

\end{document}